\documentclass{article}

\usepackage{arxiv}

\usepackage[utf8]{inputenc} 
\usepackage[T1]{fontenc}    
\usepackage{hyperref}       
\usepackage{url}            
\usepackage{booktabs}       
\usepackage{amsfonts}       
\usepackage{nicefrac}       
\usepackage{microtype}      
\usepackage{lipsum}		
\usepackage{graphicx}
\usepackage{natbib}
\usepackage[title]{appendix}
\usepackage{doi}
\usepackage{multicol}
\usepackage{orcidlink}

\title{LSC-ADL: An Activity of Daily Living (ADL)-Annotated Lifelog Dataset Generated via Semi-Automatic Clustering}

\author{
Minh-Quan Ho-Le$^{1,2}$\thanks{These authors contributed equally to this work.}\orcidlink{0009-0005-5708-7922}\\
\texttt{hlmquan@selab.hcmus.edu.vn}
\and
Duy-Khang Ho$^{1,2}$\footnotemark[1]\orcidlink{0009-0003-7584-8153}\\
\texttt{hdkhang@selab.hcmus.edu.vn}
\and
Van-Tu Ninh$^{1,2}$\orcidlink{0000-0003-0641-8806}\\
\texttt{nvtu@selab.hcmus.edu.vn}
\and
Cathal Gurrin$^{3}$\orcidlink{0000-0003-2903-3968}\\
\texttt{cathal.gurrin@dcu.ie}
\and
Minh-Triet Tran$^{1,2}$\orcidlink{0000-0003-3046-3041}\thanks{Corresponding author.}\\
\texttt{tmtriet@fit.hcmus.edu.vn}
}

\date{
{\small
$^1$Faculty of Information Technology and Software Engineering Laboratory,\\ University of Science, VNU-HCM, Ho Chi Minh City, Vietnam\\
$^2$Vietnam National University, Ho Chi Minh City, Vietnam\\
$^3$Dublin City University, Dublin, Ireland\\[1ex]
}
}

\renewcommand{\shorttitle}{\textit{arXiv} Template}

\hypersetup{
pdftitle={A template for the arxiv style},
pdfsubject={q-bio.NC, q-bio.QM},
pdfauthor={David S.~Hippocampus, Elias D.~Striatum},
pdfkeywords={First keyword, Second keyword, More},
}

\begin{document}
\maketitle

\begin{abstract}
    Lifelogging involves continuously capturing personal data through wearable cameras, providing an egocentric view of daily activities. Lifelog retrieval aims to search and retrieve relevant moments from this data, yet existing methods largely overlook activity-level annotations, which capture temporal relationships and enrich semantic understanding. In this work, we introduce LSC-ADL, an ADL-annotated lifelog dataset derived from the LSC dataset, incorporating Activities of Daily Living (ADLs) as a structured semantic layer. We generate accurate ADL annotations to enhance retrieval explainability using a semi-automatic approach featuring the HDBSCAN algorithm for intra-class clustering and human-in-the-loop verification. By integrating action recognition into lifelog retrieval, LSC-ADL bridges a critical gap in existing research, offering a more context-aware representation of daily life. We believe this dataset will advance research in lifelog retrieval, activity recognition, and egocentric vision, ultimately improving the accuracy and interpretability of retrieved content. The ADL annotations can be downloaded at 
    \url{https://bit.ly/lsc-adl-annotations}.
\end{abstract}


\section{Introduction}


Lifelogging refers to collecting and recording data about one’s everyday activities. Lifelog retrieval is searching personal lifelog data based on specific queries, aiming to provide users with relevant moments from their daily lives. In the context of this work, lifelog data primarily consists of images captured continuously by wearable cameras, offering a first-person, egocentric view of the user's life.

To improve retrieval performance, various techniques, such as query expansion, feature extraction, and similarity learning, have been explored to enhance accuracy and explainability. Among these, activity-level information—particularly Activities of Daily Living (ADLs), which refers to daily routine activities that people perform—remains an underutilized yet valuable source. Unlike static visual features, activity-level cues capture temporal relationships between frames, offering a more semantically rich understanding of the lifelogger's behaviors. This is especially important for lifelog data because ADL labels can offer helpful insights into an individual's lifestyle and behavior patterns. Since these patterns are repetitive, recognizing them can help build a more personalized and context-aware representation of daily life. 
Therefore, this presents a key opportunity for research.

In this work, we present a new dataset called LSC-ADL, which is derived from a subset of the original Lifelog Search Challenge's (LSC) dataset \cite{lsc22}, enriched with additional ADL annotations for each image. These annotations were generated using a clustering-based labeling approach with human-in-the-loop verification. We believe that the newly available dataset would encourage lifelog researchers to validate our hypothesis that this metadata could enhance the lifelog retrieval systems' performance and increase the retrieval results' explainability.

\section{Related work}

\subsection{Action recognition}
In recent years, action recognition has seen significant advancements, particularly with integrating deep learning techniques. \cite{wang2023molo} has employed the concept of motion-augmented long-short contrastive learning, which enhances few-shot action recognition by embedding motion dynamics and long-range temporal context. Additionally, \cite{zhou2023can} has applied an object-aware attention module that enriches the feature representation with object information and improves overall action recognition accuracy. \cite{wang2023actionclip} has showcased the effectiveness of multimodal learning by adapting language-image models, highlighting the potential of textual information in understanding actions. The ongoing development of intelligent human action recognition techniques continues to address challenges such as variability, complexity, and imperfect data, paving the way for more robust and efficient systems \cite{kumar2024survey}.

\subsection{Action recognition on egocentric lifelog data}
Egocentric lifelog data, collected from a first-person perspective through wearable devices and smartphones, provides a rich source of multimodal information for human activity recognition. Researchers have developed hybrid fusion frameworks that combine late and intermediate fusion mechanisms to enhance recognition performance \cite{oh2024multi}. Additionally, \cite{nie2024activity} has utilised many machine learning techniques, such as Decision Tree, K-Nearest Neighbors, Logistic Regression, etc. to recognize activity patterns. On the other hand, \cite{nguyen2018recognition} has explored the utility of hand poses and object interactions to improve recognition accuracy.
These methods aim to improve the precision and relevance of activity recognition, further enhancing the retrieval applications.

\section{Methodology}
    \subsection{Original dataset}
    The original dataset \cite{lsc22} is a multimodal lifelogging dataset collected by a single lifelogger every day over 18 months and is publicly available for research. It consists of: (1) a core image dataset containing wearable camera images captured with a narrative clip device at 1024 × 768 resolution, fully redacted to remove faces, text, and sensitive scenes for privacy compliance; (2)  metadata files providing timestamps, geographic coordinates, and semantic location descriptions; and (3) visual concepts extracted from the non-redacted version, offering object and scene information. The total number of images in the dataset is over 700,000, making it a valuable resource for lifelog research, especially lifelong retrieval. It has been used in 4 consecutive editions of the challenge from 2022 to 2025.

    To construct the LSC-ADL dataset, we selected approximately 75\% of the original images. Each image is annotated with a single activity label, informed by both the associated metadata (such as timestamps, GPS coordinates, and semantic locations) and the extracted visual features. While all available information contributes to the annotation process, the visual features play a pivotal role in determining the most representative activity for each image.

    \subsection{Label selection}
    \begin{figure}[hbt!]
	\centering
	\includegraphics[width=0.9\textwidth]{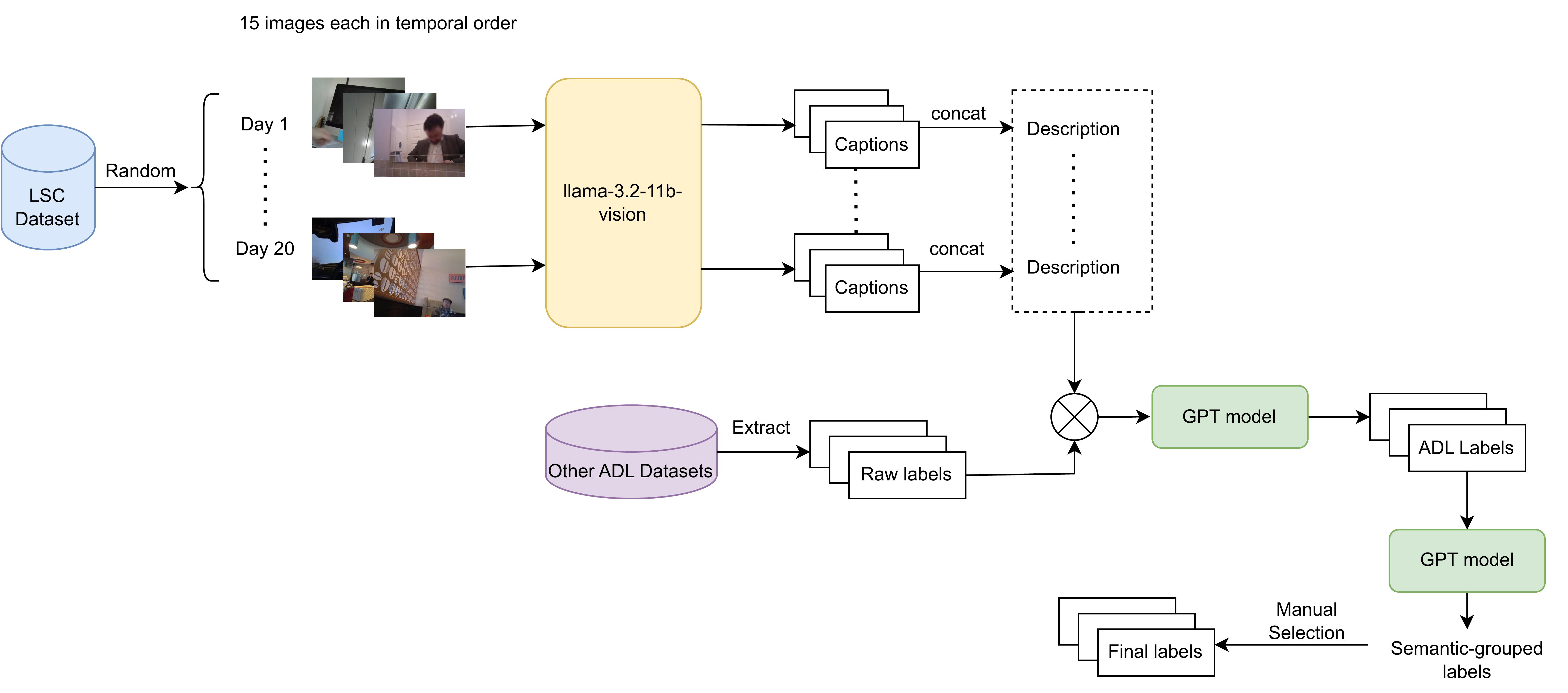}
	\caption{Label Selection Pipeline – The process begins with a random sampling of 20 days x 15 temporally ordered images from the LSC dataset. These images are processed by the LLaMA-3.2-11b-vision model to generate captions, which are concatenated into higher-level descriptions. Simultaneously, raw ADL labels are extracted from other ADL-related works. Both the generated descriptions and raw labels are fed into a GPT model for new label generation and categorization before further manual processing into final ADL classes.}
	\label{fig:LabelSelection}
    \end{figure}
    
    We developed a systematic label selection pipeline that combines insights from existing ADL datasets with a semi-automated selection process based on MLLM models to construct a well-structured set of activity labels for our dataset. This pipeline ensures that the final label set is both comprehensive and well-adapted to the LSC dataset.

    Firstly, raw activity labels are collected from four previous works \cite{nie2024activity}, \cite{nguyen2018recognition}, \cite{Cartas2017RecognizingAO}, \cite{wang2013lifelog}. These papers suggest a diverse range of daily activities, covering various aspects of human behavior. Compiling the labels from these sources creates an initial pool of candidate activities that serves as a foundation for our selection process. Since different datasets define and categorize ADLs in different ways accordingly to the individuals' lifestyles, the collected labels required further processing to ensure consistency and relevance to LSC dataset.
    
    Next, to make the action labels more aligned with the LSC dataset's lifestyle, 20 days are randomly selected. We extracted 15 images chronologically for each selected day to capture a representative sequence of activities throughout the day. Subsequently, we utilize Llama-3.2-11b-Vision \cite{llama32modelcard} to produce egocentric captions that emphasize the actions performed by the user in each frame. The generated captions were then concatenated into a single descriptive paragraph summarizing that day's activities.
    
    With these daily descriptions in hand, a GPT-based approach is employed to propose new activity labels from the previously collected raw labels. Specifically, we provided GPT with both the raw labels from other datasets and the descriptive paragraphs generated for each day. GPT was then instructed to identify which ADLs were present based on contextual relevance. This step helped filter out unnecessary labels and propose new activities meaningfully aligned with LSC's lifelog data.
    
    After extracting ADL labels for multiple days, we observed overlap and over-granularity among certain activity names. To address that problem, we removed duplicates and used GPT again to categorize the remaining labels into broader activity groups. These groups include \textit{ General}, \textit{Transport}, \textit{Food \& Beverage},\textit{Entertainment}, \textit{Work}, \textit{House Chores}, \textit{Shopping}, and \textit{Personal Hygiene}, covering a wide range of human activities. Finally, we manually reviewed the grouped labels and refined the groups to compensate for the GPT's misclassification.
    

    After correctly categorizing all labels, another manual refinement process was conducted to enhance the coherence and representativeness of the final annotation scheme. This process involved merging the overly fine-grained activities of each group into coarser classes, selecting the most representative activities within each group, and ensuring that the final set of ADL classes comprehensively captured the diversity of daily activities present in the LSC dataset. This iterative refinement established a structured and well-suited set of 35 distinct ADL classes, providing a well-defined framework for lifelog annotation and facilitating more effective activity recognition and retrieval.

    \subsection{Label proposal pipeline}

    \begin{figure}[hbt!]
	\centering
	\includegraphics[width=0.9\textwidth]{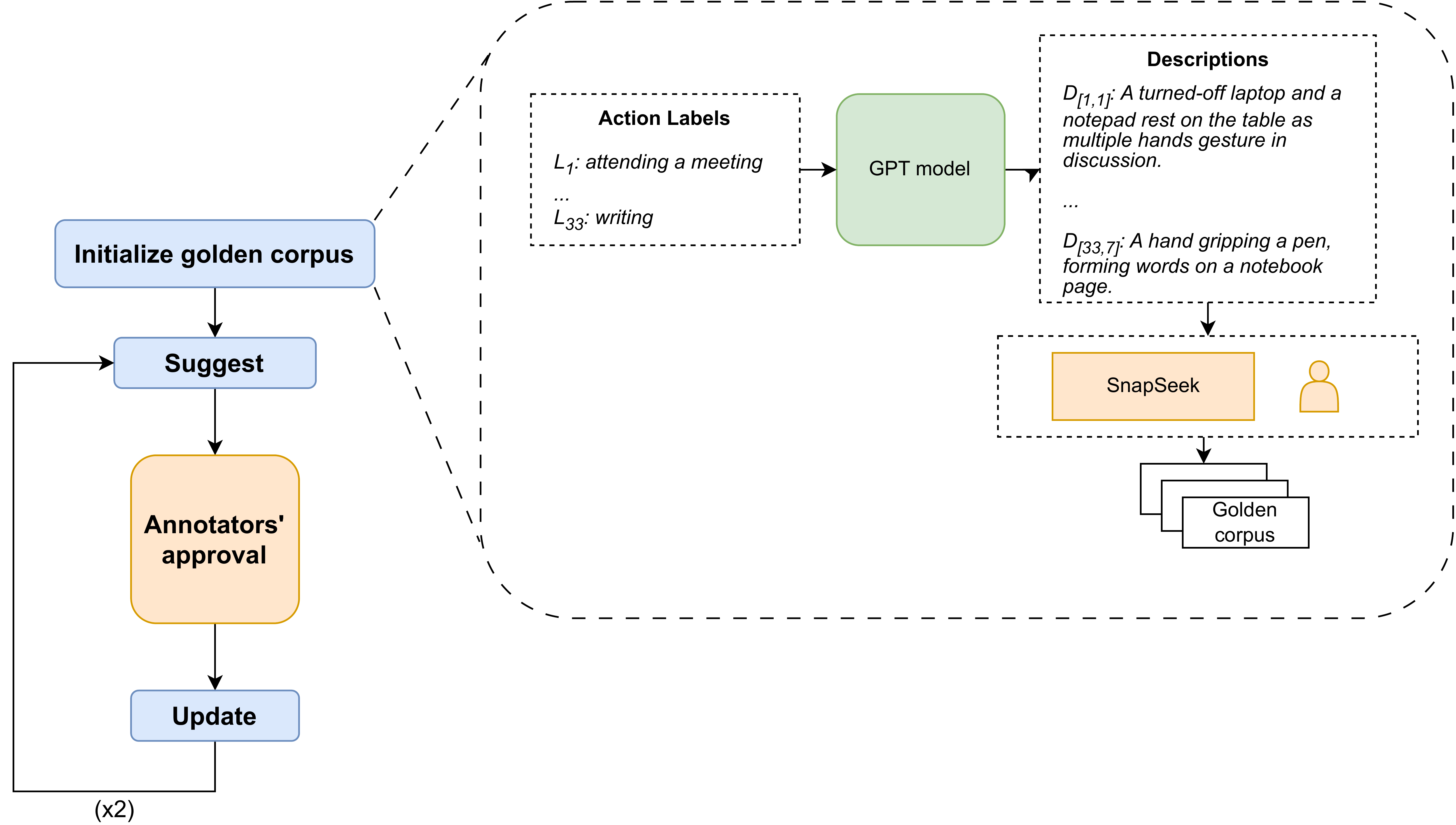}
	\caption{Label Proposal Pipeline – The process begins with initializing the golden corpus using Decompse-Expand prompting technique \cite{houenhancing} and SnapSeek retrieval system \cite{snapseek}, followed by generating label suggestions. Annotators review the suggestions and approve or modify them before re-clustering and updating the corpus iteratively.}
	\label{fig:LabelProposal}
    \end{figure}
    
    Using the predefined list of ADL classes above, we propose a pipeline that features both machine learning and human supervision in an interleaving manner. Our approach includes a clustering algorithm to group relevant images and map each cluster to one specific ADL class. By gradually adding more images to the approved list, we expand the cluster's border, or even create new clusters to scan all images that may belong to different variances of one ADL class.

    Fig. \ref{fig:LabelProposal} demonstrates four stages of the pipeline.
    \begin{enumerate}

        \item \textbf{Stage 1: Golden corpus initialization}. For each ADL class, we utilize the Decompose-Expand prompting technique \cite{houenhancing} to generate seven descriptive sentences that reflect the lifelogger’s perspective. To briefly describe the technique, for each class, we ask GPT to generate object and action keywords that may appear in the egocentric images of that activity. Subsequently, we request the model to generate possible egocentric descriptions based on the classes' names and keywords. These descriptive sentences are then reviewed and paraphrased to better align with the dataset. The resulting descriptions can be viewed in Appendix \ref{appendix:labeldescript}. With each refined description, we employ the SnapSeek retrieval system \cite{snapseek} to retrieve relevant images with the LSC dataset as the image source. This process yields approximately 300 initial seed samples per class.
        
        \item \textbf{Stage 2: Label suggestion}. We compute a representative vector for each cluster within every class by averaging the vectors of all images in the cluster. For every unannotated image, we calculate its cosine similarity with these class's representations and assign a label based on the highest similarity score across the 35 ADL classes. A similarity threshold is then applied to filter out uncertain predictions, retaining only the images with scores above this threshold.
        \item \textbf{Stage 3: Human approval}. In this phase, human annotators review the labels suggested in the previous step. We recruited 20 participants, each tasked with confirming or rejecting the suggested labels. Their decisions are collected and used to refine the clustering in the subsequent step. Annotators follow a predefined set of labeling rules, which include specific guidelines for each ADL class. This rule set is continuously refined based on annotator feedback across iterations.
        \item \textbf{Stage 4: Cluster update}. After we approve new images, we add them to their matching class in the dataset. We believe that each activity, i.e. ADL class, 
         might occur in different scenarios, even though they belong to the same class. To better capture these variations, we re-cluster the images within each class using the HDBSCAN algorithm. HDBSCAN is good at finding groups of closely related images, even if those groups vary in size or density. In our case, each cluster within a class might represent a different scenario of that activity. Therefore, instead of representing an entire class with a single average feature vector, we utilize multiple average vectors corresponding to different clusters within each class. This helps to understand the differences within a class better and make more accurate predictions. 
        After this step, the process goes back to Stage 2, and we repeat the whole cycle twice to keep improving and growing the labeled dataset.
        
    \end{enumerate}

\section{Exploratory data analysis}
    In this section, we briefly analyze the newly generated activity labels, both in isolation and in relation to other data fields.

    \begin{figure}[hbt!] \centering \includegraphics[width=0.8\textwidth]{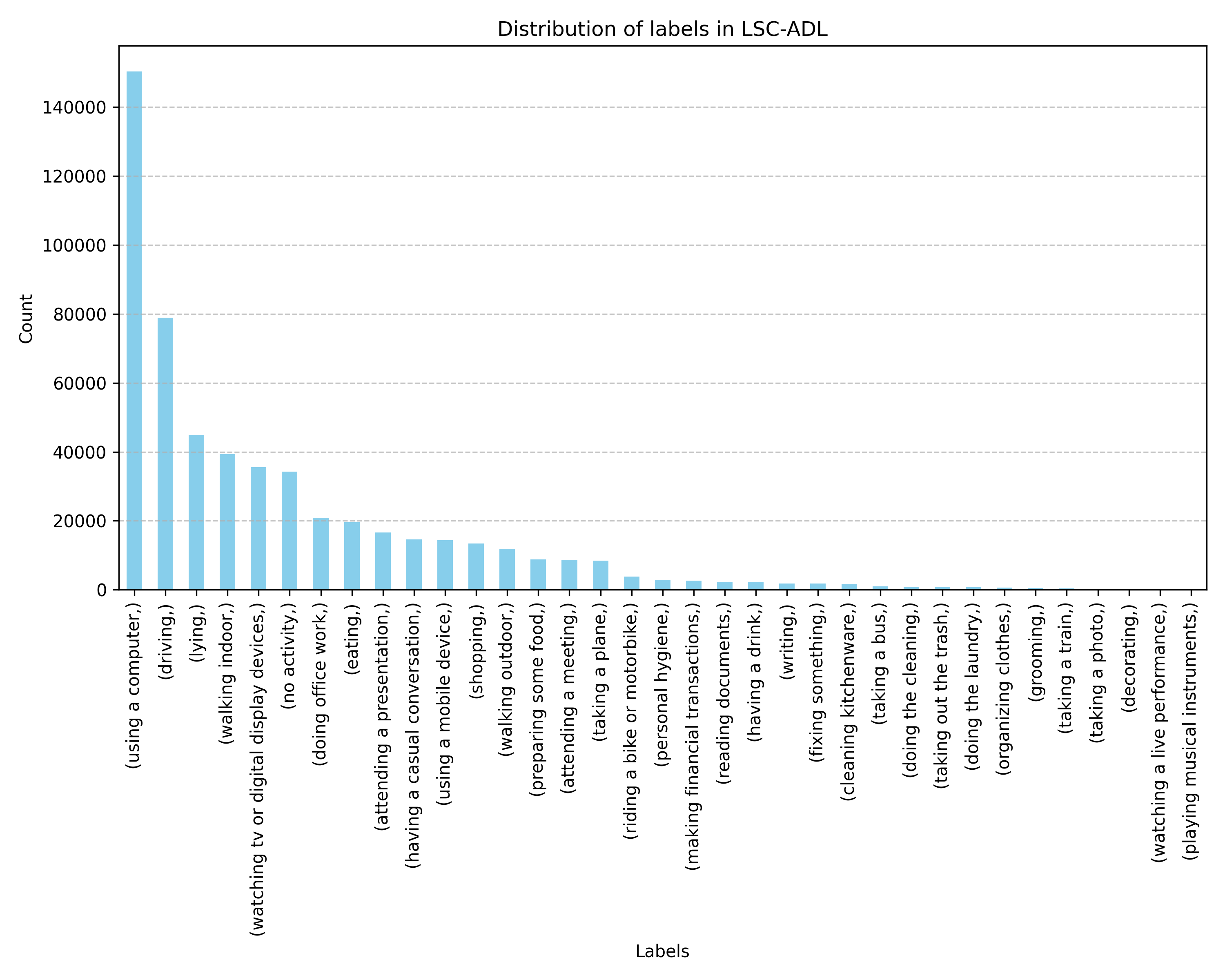} \caption{Label distribution across the dataset.} \label{fig:LabelDistribution} \end{figure}
    
    As illustrated in Fig. \ref{fig:LabelDistribution}, the dataset exhibits a pronounced long-tailed distribution. Roughly 80\% of the labeled samples belong to just 10 out of the 35 activity classes. These dominant classes primarily consist of common Activities of Daily Living (ADLs) such as driving, using a computer, doing office work, and eating—activities frequently observed in everyday human routines. Notably, the most represented class, using a computer, contains nearly 150,000 instances. In contrast, 11 classes have fewer than 1,000 samples each, forming the sparse and extended tail of the distribution. These rare classes represent highly specific actions or uncommon scenarios the lifelogger does not regularly encounter. This highlights the personal nature of the data and the challenge of capturing a truly comprehensive activity set in real-world lifelogging contexts.

    \begin{figure}[hbt!] \centering \includegraphics[width=0.8\textwidth]{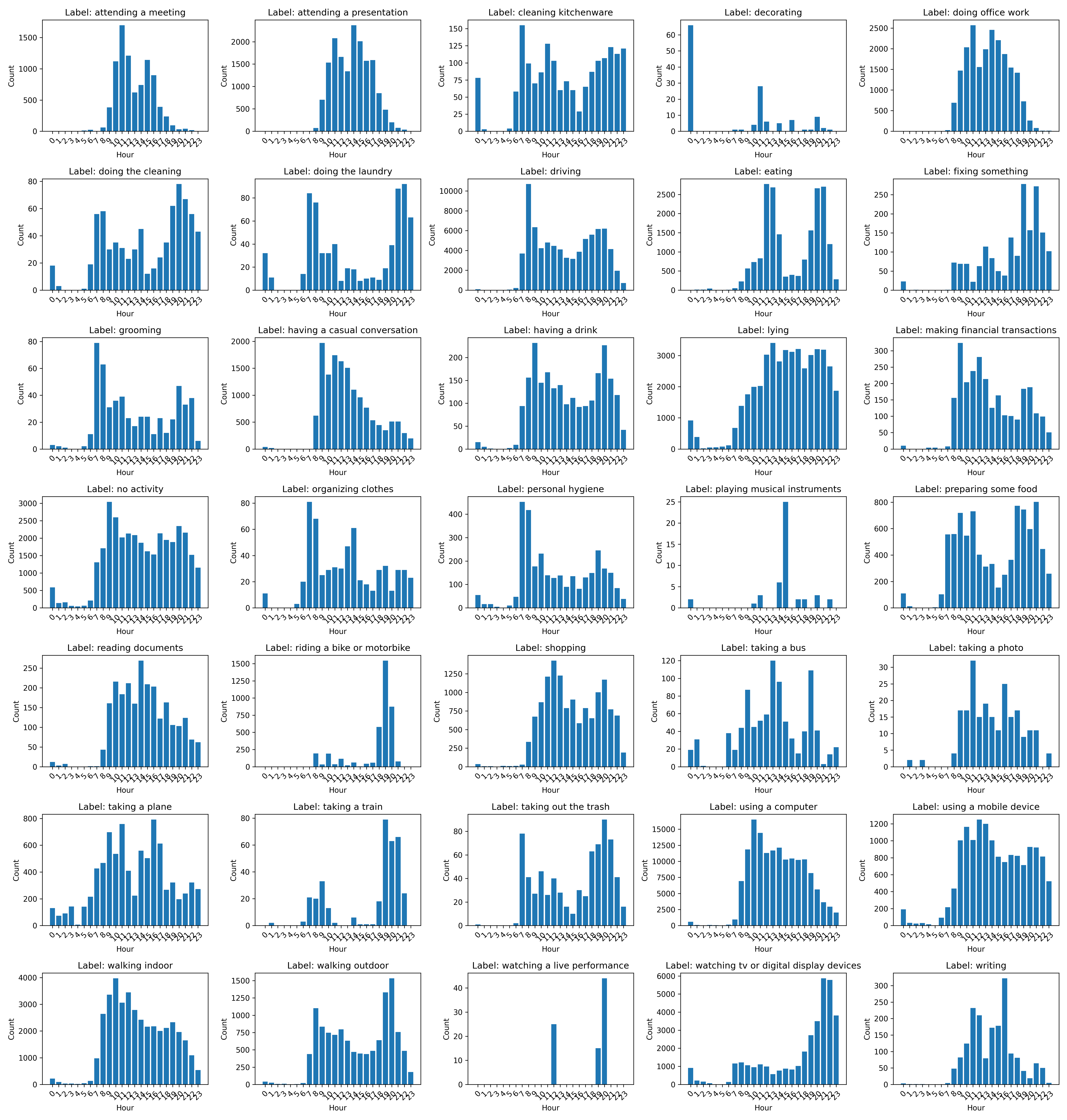} \caption{Hour distribution per label.} \label{fig:HourDistribution} \end{figure}

    Fig. \ref{fig:HourDistribution} presents another layer of insight into the dataset by illustrating the temporal patterns of daily activities. These hourly distributions reveal meaningful routines aligned with common daily life. For instance, attending a meeting, attending a presentation, and doing office work all show a clear concentration of activity during the daytime, with two noticeable peaks—one in the morning and one in the afternoon—and a dip around noon, likely corresponding to lunch breaks. This pattern reflects a structured office environment. Other classes display distinct time-based behaviors. For example, eating shows two strong peaks: around midday (lunch) and late evening (dinner). Personal hygiene activities predominantly occur in the early morning hours as part of the start-of-day routine. In contrast, watching TV or digital displays tends to spike in the evening, aligning with relaxation time at home after a working day.

\section{Conclusion}
    In conclusion, our proposed LSC-ADL dataset introduces a valuable resource for advancing lifelog retrieval by incorporating activity-level annotations. By leveraging ADL labels, we aim to enhance both the accuracy and explainability of retrieval systems, providing a more semantically rich representation of daily activities. Our clustering-based labeling approach, combined with human-in-the-loop verification, ensures reliable annotations that can support further research in lifelogging. We hope this dataset will encourage the community to explore the role of activity understanding in lifelog retrieval and inspire the development of more personalized and context-aware retrieval models.

\section{Acknowledgements}
    We would like to express our sincere gratitude to the Software Engineering Laboratory (SELAB) at the University of Science, VNU-HCM (HCMUS) for providing the computing resources and infrastructure that made this research possible. We also appreciate the collaborators from the Faculty of Information Technology at HCMUS (FIT@HCMUS) and Hanoi University of Science and Technology (HUST) for their valuable support in giving human approval for the label proposal pipeline.

\bibliographystyle{unsrtnat}
\bibliography{references}  

\begin{thebibliography}{13}
\providecommand{\natexlab}[1]{#1}
\providecommand{\url}[1]{\texttt{#1}}
\expandafter\ifx\csname urlstyle\endcsname\relax
  \providecommand{\doi}[1]{doi: #1}\else
  \providecommand{\doi}{doi: \begingroup \urlstyle{rm}\Url}\fi

\bibitem[Gurrin et~al.(2022)Gurrin, Zhou, Healy, \TH{}\'{o}r~J\'{o}nsson, Dang-Nguyen, Loko\'{c}, Tran, H\"{u}rst, Rossetto, and Sch\"{o}ffmann]{lsc22}
Cathal Gurrin, Liting Zhou, Graham Healy, Bj\"{o}rn \TH{}\'{o}r~J\'{o}nsson, Duc-Tien Dang-Nguyen, Jakub Loko\'{c}, Minh-Triet Tran, Wolfgang H\"{u}rst, Luca Rossetto, and Klaus Sch\"{o}ffmann.
\newblock Introduction to the fifth annual lifelog search challenge, lsc'22.
\newblock In \emph{Proceedings of the 2022 International Conference on Multimedia Retrieval}, ICMR '22, page 685–687, New York, NY, USA, 2022. Association for Computing Machinery.
\newblock ISBN 9781450392389.
\newblock \doi{10.1145/3512527.3531439}.
\newblock URL \url{https://doi.org/10.1145/3512527.3531439}.

\bibitem[Wang et~al.(2023{\natexlab{a}})Wang, Zhang, Qing, Gao, Zhang, Zhao, and Sang]{wang2023molo}
Xiang Wang, Shiwei Zhang, Zhiwu Qing, Changxin Gao, Yingya Zhang, Deli Zhao, and Nong Sang.
\newblock Molo: Motion-augmented long-short contrastive learning for few-shot action recognition.
\newblock In \emph{Proceedings of the IEEE/CVF conference on computer vision and pattern recognition}, pages 18011--18021, 2023{\natexlab{a}}.

\bibitem[Zhou et~al.(2023)Zhou, Arnab, Sun, and Schmid]{zhou2023can}
Xingyi Zhou, Anurag Arnab, Chen Sun, and Cordelia Schmid.
\newblock How can objects help action recognition?
\newblock In \emph{Proceedings of the IEEE/CVF Conference on Computer Vision and Pattern Recognition}, pages 2353--2362, 2023.

\bibitem[Wang et~al.(2023{\natexlab{b}})Wang, Xing, Mei, Liu, and Jiang]{wang2023actionclip}
Mengmeng Wang, Jiazheng Xing, Jianbiao Mei, Yong Liu, and Yunliang Jiang.
\newblock Actionclip: Adapting language-image pretrained models for video action recognition.
\newblock \emph{IEEE Transactions on Neural Networks and Learning Systems}, 2023{\natexlab{b}}.

\bibitem[Kumar and Kumar(2024)]{kumar2024survey}
Rahul Kumar and Shailender Kumar.
\newblock A survey on intelligent human action recognition techniques.
\newblock \emph{Multimedia Tools and Applications}, 83\penalty0 (17):\penalty0 52653--52709, 2024.

\bibitem[Oh and Kim(2024)]{oh2024multi}
YongKyung Oh and Sungil Kim.
\newblock Multi-modal lifelog data fusion for improved human activity recognition: A hybrid approach.
\newblock \emph{Information Fusion}, 110:\penalty0 102464, 2024.

\bibitem[Nie et~al.(2024)Nie, Gurrin, and Scriney]{nie2024activity}
Dongyun Nie, Cathal Gurrin, and Michael Scriney.
\newblock Activity classification for daily lifelogs.
\newblock In \emph{Proceedings of the 1st ACM Workshop on AI-Powered Q\&A Systems for Multimedia}, pages 31--35, 2024.

\bibitem[Nguyen et~al.(2018)Nguyen, Nebel, and Florez-Revuelta]{nguyen2018recognition}
Thi-Hoa-Cuc Nguyen, Jean-Christophe Nebel, and Francisco Florez-Revuelta.
\newblock Recognition of activities of daily living from egocentric videos using hands detected by a deep convolutional network.
\newblock In \emph{Image Analysis and Recognition: 15th International Conference, ICIAR 2018, P{\'o}voa de Varzim, Portugal, June 27--29, 2018, Proceedings 15}, pages 390--398. Springer, 2018.

\bibitem[Cartas et~al.(2017)Cartas, Mar{\'i}n, Radeva, and Dimiccoli]{Cartas2017RecognizingAO}
Alejandro Cartas, Juan Mar{\'i}n, Petia Radeva, and Mariella Dimiccoli.
\newblock Recognizing activities of daily living from egocentric images.
\newblock In \emph{Iberian Conference on Pattern Recognition and Image Analysis}, 2017.
\newblock URL \url{https://api.semanticscholar.org/CorpusID:7057654}.

\bibitem[Wang and Smeaton(2013)]{wang2013lifelog}
Peng Wang and Alan Smeaton.
\newblock Using visual lifelogs to automatically characterize everyday activities.
\newblock \emph{Information Sciences}, 230:\penalty0 147–161, 05 2013.
\newblock \doi{10.1016/j.ins.2012.12.028}.

\bibitem[AI@Meta(2024)]{llama32modelcard}
AI@Meta.
\newblock Llama 3.2 vision model card.
\newblock 2024.
\newblock URL \url{https://github.com/meta-llama/llama-models/blob/main/models/llama3_2/MODEL_CARD_VISION.md}.

\bibitem[Hou et~al.()Hou, Zhang, Li, Wang, and Huang]{houenhancing}
Puyue Hou, Jinjin Zhang, Guohao Li, Guodong Wang, and Di~Huang.
\newblock Enhancing video understanding with vision and language collaboration.

\bibitem[Ho-Le et~al.(2024)Ho-Le, Do-Huu, Ho, Le-Hinh, Vo-Hoang, Ninh, and Tran]{snapseek}
Minh-Quan Ho-Le, Huy-Hoang Do-Huu, Duy-Khang Ho, Nhut-Thanh Le-Hinh, Hoa-Vien Vo-Hoang, Van-Tu Ninh, and Minh-Triet Tran.
\newblock Snapseek: An interactive lifelog acquisition system for lsc'24.
\newblock In \emph{Proceedings of the 7th Annual ACM Workshop on the Lifelog Search Challenge}, LSC '24, page 24–29, New York, NY, USA, 2024. Association for Computing Machinery.
\newblock ISBN 9798400705502.
\newblock \doi{10.1145/3643489.3661116}.
\newblock URL \url{https://doi.org/10.1145/3643489.3661116}.

\end{thebibliography}

\begin{appendices}
\section{Labels and Descriptions} \label{appendix:labeldescript}

\begin{multicols}{2}
\begin{enumerate}

    \item \textbf{Attending a Meeting}
    \begin{itemize}
        \item A conference table with colleagues seated around, notebooks and pens scattered across the surface.
        \item A blurred hand gesture from a speaking colleague while a printed agenda rests on the table.
        \item A cup of coffee near a laptop, with a projector screen displaying a PowerPoint slide.
        \item A view of name tags and folders on the table as a colleague takes notes.
        \item A whiteboard covered with diagrams and keywords, with someone pointing at it.
        \item A wristwatch in the professor’s peripheral view as he listens to a discussion.
        \item A blurry movement of people shaking hands at the end of the meeting.
    \end{itemize}

    \item \textbf{Attending a Presentation}
    \begin{itemize}
        \item A large screen with a slide showing bullet points and a presenter gesturing.
        \item Rows of seated attendees’ backs visible in front of the professor’s view.
        \item A table with the presentation agenda and a microphone.
        \item The presenter's hands moving mid-sentence as they explain a concept.
        \item A dimly lit conference hall with spotlights on the stage.
        \item The silhouette of a speaker against a bright projection screen.
        \item A laser pointer’s red dot highlighting text on the slide.
    \end{itemize}

    \item \textbf{Cleaning Kitchenware}
    \begin{itemize}
        \item A stream of water running over a soapy plate in the sink.
        \item A dish sponge in the professor’s hand scrubbing a greasy pan.
        \item A hand placing a cleaned glass onto a drying rack.
        \item A bottle of dish soap foaming up as it’s squeezed.
        \item A sink filled with bubbles, utensils submerged underneath.
        \item A towel wiping droplets off a shiny fork.
        \item A hand arranging washed plates into a cupboard.
    \end{itemize}

    \item \textbf{Decorating}
    \begin{itemize}
        \item A plant being placed on the shelf.
        \item A string of fairy lights being adjusted along the ceiling.
        \item A hand holding a roll of tape while securing a banner.
        \item A close-up of ornaments being arranged on a shelf.
        \item A box of decorative items partially unpacked on the floor.
        \item A framed certificate being aligned on a wall.
        \item A hand smoothing out wrinkles on a tablecloth.
    \end{itemize}

    \item \textbf{Doing Office Work}
    \begin{itemize}
        \item A desk covered with printed reports.
        \item A bookshelf with stacked academic journals in view.
        \item A pen tapping against the table while thinking.
        \item A coffee cup resting next to scattered handwritten notes.
        \item A printer being used to print documents.
        \item A calendar open on the desk with upcoming meetings marked.
        \item A hand placing a stapled document into a file folder.
    \end{itemize}

    \item \textbf{Doing the Cleaning}
    \begin{itemize}
        \item A mop gliding across a wooden floor with streaks of water behind.
        \item A hand pressing a cleaning spray onto a glass window.
        \item A vacuum cleaner cord stretching across the floor.
        \item A cloth wiping dust from a wooden desk.
        \item A hand squeezing out excess water from a mop bucket.
        \item A dustpan being used to collect the dust.
        \item A broom sweeping small debris into a dustpan.
    \end{itemize}

    \item \textbf{Doing the Laundry}
    \begin{itemize}
        \item A hand pouring detergent into a washing machine drawer.
        \item Clothes being put into the washing machine.
        \item A washing machine door being closed with clothes inside.
        \item A pair of socks being hung on a drying rack.
        \item A hand collecting a newly washed shirt out of the washing machine.
        \item A laundry basket filled with dirty clothes.
        \item A dryer lint filter being cleaned out.
    \end{itemize}

    \item \textbf{Eating}
    \begin{itemize}
        \item A fork lifting a piece of food from a plate.
        \item A bowl of steaming soup in front of the professor.
        \item A spoon dipping into a dessert cup.
        \item A napkin resting next to a half-eaten sandwich.
        \item A hand holding a hamburger.
        \item A piece of bread being buttered with a knife.
        \item A plate with only crumbs remaining after a meal.
    \end{itemize}

    \item \textbf{Fixing Something}
    \begin{itemize}
        \item A screwdriver turning a loose screw on a chair.
        \item A toolbox open with scattered tools on a workbench.
        \item A hand holding a glue gun while repairing a broken object.
        \item A battery being replaced inside a remote control.
        \item A tangled cord being untwisted and organized.
        \item A watch back cover being removed to change the battery.
        \item A bent paperclip being straightened to fix a jammed stapler.
    \end{itemize}

    \item \textbf{Grooming}
    \begin{itemize}
        \item A hand adjusting a tie while looking into a mirror.
        \item A person touching his hair while looking into a mirror.
        \item A razor gliding across a foamy chin.
        \item A hand spraying cologne onto the wrist.
        \item A lint roller being used on a blazer.
        \item A cufflink being fastened onto a dress shirt.
        \item A watch being adjusted on the professor’s wrist.
    \end{itemize}

    \item \textbf{Having a Casual Conversation}
    \begin{itemize}
        \item A colleague gesturing while speaking in a hallway.
        \item A person opening his mouth as if he is talking.
        \item A slight head tilt, showing engagement in the discussion.
        \item A shared document on the table with fingers pointing at a section.
        \item A blurry hand movement emphasizing a point.
        \item A subtle smile visible from the side of a face.
        \item A handshake at the end of the conversation.
    \end{itemize}

    \item \textbf{Having a Drink}
    \begin{itemize}
        \item A coffee mug being lifted from the table.
        \item A water bottle cap being twisted open.
        \item A sip of tea leaving condensation.
        \item A can of soda being popped open.
        \item A steaming cup of coffee with a spoon resting beside it.
        \item A hand holding a wine glass at a formal gathering.
        \item A nearly empty bottle with droplets visible inside.
    \end{itemize}

    \item \textbf{Lying}
    \begin{itemize}
        \item A pillow in the professor’s field of view.
        \item A hand resting on the chest while looking up at the ceiling.
        \item A bedside table with a dimly lit lamp.
        \item A blanket partially covering the professor’s legs.
        \item A smartphone resting nearby on a bedspread.
        \item A book lying open next to the professor.
        \item A view of a ceiling fan slowly spinning overhead.
    \end{itemize}

    \item \textbf{Making Financial Transactions}
    \begin{itemize}
        \item A credit card being inserted into a payment terminal.
        \item A wallet being opened with bills inside.
        \item A hand signing a receipt with a pen.
        \item A cashier scanning items at the register.
        \item A printed receipt resting on a counter.
        \item A stack of coins next to a register.
        \item A point-of-sale screen showing a total amount due.
    \end{itemize}

    \item \textbf{No Activity}
    \begin{itemize}
        \item A blurred image of a hand moving in front of the camera.
        \item A dark, occluded view from inside a pocket.
        \item A tilted, unclear shot of an empty room.
        \item A distorted reflection in a glass surface.
        \item An obstruction by a sleeve or clothing.
        \item A sudden bright flash from an overexposed shot.
        \item A completely black or white screen due to motion blur.
    \end{itemize}

    \item \textbf{Organizing Clothes}
    \begin{itemize}
        \item A hand folding a neatly ironed shirt.
        \item A dresser drawer being pulled open, revealing stacks of clothes.
        \item A closet with hangers lined up, some shirts being adjusted.
        \item A hand placing socks into a designated section.
        \item A belt being coiled neatly and stored.
        \item A pile of clothes being sorted into categories.
        \item A hand smoothing wrinkles out of a folded sweater.
    \end{itemize}

    \item \textbf{Personal Hygiene}
    \begin{itemize}
        \item A toothbrush covered in foam brushing against teeth.
        \item A soap dispenser being pressed with a drop landing on the palm.
        \item A hand scrubbing with foamy lather.
        \item A towel dabbing water off a freshly washed face.
        \item A bottle of shampoo being squeezed onto the palm.
        \item A hand gripping a deodorant stick under an armpit.
        \item A hand flossing between teeth in front of a mirror.
    \end{itemize}

    \item \textbf{Playing Musical Instruments}
    \begin{itemize}
        \item A hand pressing keys on a piano.
        \item A guitar pick strumming against metal strings.
        \item A violin bow gliding across the strings.
        \item A page of sheet music propped up on a stand.
        \item A trumpet valve being pressed mid-note.
        \item A drumstick striking a snare drum.
        \item A hand adjusting the tuning pegs of a guitar.
    \end{itemize}

    \item \textbf{Preparing Some Food}
    \begin{itemize}
        \item A knife slicing through fresh vegetables on a cutting board.
        \item A pot of boiling water with steam rising.
        \item A hand cracking an egg into a frying pan.
        \item A spoon stirring a simmering sauce.
        \item A sandwich being assembled with layered ingredients.
        \item A blender filled with fruits being turned on.
        \item A tray of cookies being placed into an oven.
    \end{itemize}

    \item \textbf{Reading Documents}
    \begin{itemize}
        \item A hand holding a printed research paper with highlighted text.
        \item A close-up of fingers flipping through legal-sized pages.
        \item A desk cluttered with academic journals and a notepad.
        \item A finger pointing to a section of a manuscript.
        \item A thick binder opened to a specific section.
        \item A post-it note sticking out from the edge of a document.
        \item A reading lamp casting light over a dense stack of papers.
    \end{itemize}

    \item \textbf{Riding a Bike or Motorbike}
    \begin{itemize}
        \item A view of handlebars and a front wheel rolling down a path.
        \item A hand gripping the throttle of a motorbike.
        \item A speedometer showing the current pace.
        \item A helmet strap being fastened under the chin.
        \item A road stretching ahead with motion blur.
        \item A reflection of the professor in a bike’s side mirror.
        \item A bike tire being visible in the view.
    \end{itemize}

    \item \textbf{Shopping}
    \begin{itemize}
        \item A hand reaching for a product on a supermarket shelf.
        \item A shopping cart filled with groceries.
        \item A price tag being examined up close.
        \item A salesman introducing a product.
        \item A mannequin wearing fashionable clothes.
        \item Aisles in the supermarket with products.
        \item A hand pulling a reusable bag open.
    \end{itemize}

    \item \textbf{Taking a Bus}
    \begin{itemize}
        \item A ticket being inserted into a validation machine.
        \item A view of a crowded aisle from a seated perspective.
        \item A bus stop sign visible through the window.
        \item A hand gripping a safety rail overhead.
        \item A digital display showing the next stop.
        \item A backpack resting on the professor’s lap.
        \item A bus driver visible in the front mirror.
    \end{itemize}

    \item \textbf{Taking a Car}
    \begin{itemize}
        \item A seatbelt being fastened across the chest.
        \item A GPS screen displaying directions.
        \item A hand adjusting the rearview mirror.
        \item A car key being inserted into the ignition.
        \item A hand gripping the steering wheel.
        \item A view through the windshield at a red light.
        \item A dashboard showing fuel and speed indicators.
    \end{itemize}
    \item \textbf{Taking a Photo}
    \begin{itemize}
        \item A smartphone held up with a camera interface on screen.
        \item A hand pressing the shutter button.
        \item A blurry hand adjusting focus on a DSLR lens.
        \item A subject posing in the frame.
        \item A reflection in a camera lens.
        \item A photo preview on the camera’s LCD screen.
        \item A tripod being set up in an indoor space.
    \end{itemize}

    \item \textbf{Taking a Plane}
    \begin{itemize}
        \item A turned-off screen of the TV at the back of the plane's seat.
        \item A row of airplane seats with passengers settling in.
        \item A view of the wing through a small window.
        \item A seatbelt being fastened before takeoff.
        \item A tray table unfolded with a meal on it.
        \item A flight attendant passing by with a beverage cart.
        \item An overhead compartment being opened to store luggage.
    \end{itemize}

    \item \textbf{Taking a Train}
    \begin{itemize}
        \item A train ticket being scanned at the station entrance.
        \item A window seat view showing the moving landscape.
        \item A hand holding onto a vertical pole while standing.
        \item A digital sign displaying the next station name.
        \item A coffee cup resting on a foldable tray.
        \item A hand gripping a rolling suitcase handle.
        \item A reflection of the professor in the train window.
    \end{itemize}

    \item \textbf{Taking Out the Trash}
    \begin{itemize}
        \item A black trash bag being lifted from a bin.
        \item A hand tying the top of a full garbage bag.
        \item A bin lid being opened with trash visible inside.
        \item A recycling bin next to a general waste bin.
        \item A sidewalk with lined-up garbage bags.
        \item A gloved hand carrying a bag to the curb.
        \item A trash chute door being pulled open.
    \end{itemize}

    \item \textbf{Using a Computer}
    \begin{itemize}
        \item A keyboard with fingers typing mid-sentence.
        \item A computer screen displaying lines of code.
        \item A mouse being clicked next to a notebook.
        \item A video call window open on a laptop screen.
        \item A document with highlighted text and comments.
        \item A taskbar with multiple applications running.
        \item A coffee cup beside an open laptop.
    \end{itemize}

    \item \textbf{Using a Mobile Device}
    \begin{itemize}
        \item A smartphone screen displaying a messaging app.
        \item A hand scrolling through social media.
        \item A video playing in fullscreen mode.
        \item A thumb tapping on a virtual keyboard.
        \item A call interface with an ongoing conversation.
        \item A phone being plugged into a charging cable.
        \item A weather app showing today’s forecast.
    \end{itemize}

    \item \textbf{Walking Indoor}
    \begin{itemize}
        \item A hallway stretching ahead with doors on both sides.
        \item A staircase with a hand gripping the railing.
        \item A conference room door being pushed open.
        \item A tiled floor reflecting overhead lights.
        \item A wristwatch in view as the professor walks.
        \item A coat draped over an arm while moving through a corridor.
        \item A set of elevators with buttons being pressed.
    \end{itemize}

    \item \textbf{Walking Outdoor}
    \begin{itemize}
        \item A sidewalk lined with trees on both sides.
        \item A crosswalk signal blinking green.
        \item A coffee cup being carried while walking.
        \item A bag strap over the professor’s shoulder in view.
        \item A reflection of movement in a store window.
        \item A parked bicycle visible on the curbside.
        \item A distant building entrance coming into focus.
    \end{itemize}

    \item \textbf{Watching a Live Performance}
    \begin{itemize}
        \item A dimly lit stage with spotlights on performers.
        \item A hand holding a concert program booklet.
        \item A large crowd seated in a theater.
        \item A choir singing in the church.
        \item A performer mid-action with dramatic lighting.
        \item A blurred view of clapping hands.
        \item A ticket stub resting on a lap.
    \end{itemize}

    \item \textbf{Watching TV or Digital Display Devices}
    \begin{itemize}
        \item A remote control being held towards a large screen.
        \item An advertisement display device in a public area.
        \item A dimly lit room with a glowing TV.
        \item A news ticker scrolling across a display.
        \item A hand adjusting the volume button.
        \item A reflection of the screen on a pair of glasses.
        \item A streaming service interface with movie options.
    \end{itemize}

    \item \textbf{Writing}
    \begin{itemize}
        \item A pen gliding across a lined notebook.
        \item A checklist being marked with a bold black pen, one box at a time.
        \item A clipboard with forms being filled out in neat handwriting.
        \item A highlighter highlighting text on a paper.
        \item A fountain pen refilling from an ink bottle.
        \item An electronic pen touching a tablet.
        \item A signature being added at the bottom of a document.
    \end{itemize}

\end{enumerate}
\end{multicols}
\end{appendices}






\end{document}